\documentclass[10pt,twocolumn,letterpaper]{article}

\usepackage{cvpr}
\usepackage{times}
\usepackage{epsfig}
\usepackage{graphicx}
\usepackage{amsmath}
\usepackage{amssymb}

\usepackage[pagebackref=true,breaklinks=true,letterpaper=true,colorlinks,bookmarks=false]{hyperref}

\usepackage{times}
\usepackage{epsfig}
\usepackage{graphicx}
\usepackage{amsmath}
\usepackage{amssymb}
\usepackage{pifont}

\newcommand*\colorcmark[1]{%
  \expandafter\newcommand\csname #1cmark\endcsname{\textcolor{#1}{\ding{51}}}%
}
\colorcmark{green}

\newcommand*\colorxmark[1]{%
  \expandafter\newcommand\csname #1xmark\endcsname{\textcolor{#1}{\ding{55}}}%
}
\colorxmark{red}

\usepackage{multirow}
\usepackage{enumitem}
\usepackage{subcaption}
\usepackage[table]{xcolor}
\usepackage{amsthm}

\usepackage{xspace}

\definecolor{lightblue}{rgb}{0.93,0.95,1.0}

 \cvprfinalcopy 

\makeatletter
\DeclareRobustCommand\onedot{\futurelet\@let@token\@onedot}
\def\@onedot{\ifx\@let@token.\else.\null\fi\xspace}

\def\eg{\emph{e.g}\onedot} 
\def\ie{\emph{i.e}\onedot} 
 
\def\etc{\emph{etc}\onedot} 
\def\wrt{w.r.t\onedot} 
\def\etal{\emph{et al}\onedot}
\makeatother

\newlength\savewidth\newcommand\shline{\noalign{\global\savewidth\arrayrulewidth
  \global\arrayrulewidth 1pt}\hline\noalign{\global\arrayrulewidth\savewidth}}

\newenvironment{myitemize}
{ \begin{itemize}  [leftmargin=*]
    \setlength{\itemsep}{0pt}
    \setlength{\parskip}{0pt}
    \setlength{\parsep}{0pt}}
{ \end{itemize} }

\ifcvprfinal\fi
\begin{document}

\title{Video Modeling with Correlation Networks}

\author{Heng Wang \qquad Du Tran \qquad Lorenzo Torresani  \qquad Matt Feiszli\\
Facebook AI\\
{\tt\small \{hengwang,trandu,torresani,mdf\}@fb.com}
}

\maketitle

\begin{abstract}
Motion is a salient cue to recognize actions in video. Modern action recognition models leverage motion information either explicitly by using optical flow as input or implicitly by means of 3D convolutional filters that simultaneously capture appearance and motion information. This paper proposes an alternative approach based on a learnable correlation operator that can be used to establish frame-to-frame matches over convolutional feature maps in the different layers of the network. The proposed architecture enables the fusion of this explicit temporal matching information with traditional appearance cues captured by 2D convolution. Our correlation network compares favorably with widely-used 3D CNNs for video modeling, and achieves competitive results over the prominent two-stream network while being much faster to train. We empirically demonstrate that correlation networks produce strong results on a variety of video datasets, and outperform the state of the art on four popular benchmarks for action recognition: Kinetics, Something-Something, Diving48 and Sports1M.
\end{abstract}

\section{Introduction}

After the breakthrough of AlexNet~\cite{Krizhevsky12} on ImageNet~\cite{deng2009imagenet}, convolutional neural networks (CNNs) have become the dominant model for still-image classification~\cite{lecun1998gradient,SimonyanZ14a,Szegedy15,KaimingHe16}. 
In the video domain, CNNs were initially adopted as image-based feature extractor on individual frames of the video~\cite{Karpathy14}. More recently, CNNs for video analysis have been extended with the capability of capturing not only  appearance information contained in individual frames but also motion information extracted from the temporal dimension of the image sequence. This is usually achieved by one of two possible mechanisms. One strategy involves the use of a two-stream network~\cite{SimonyanZ14,wang2015towards,FeichtenhoferPZ16,WangXW0LTG16,qiu2019learning,crasto2019mars} where one stream operates on RGB frames to model appearance information and the other stream extracts motion features from optical flow provided as input. The representations obtained from these two distinct inputs are then fused, typically in a late layer of the network. An alternative strategy is to use 3D convolutions~\cite{baccouche2011sequential,Ming2013,Tran15,linsuniccv15,tran2017closer,P3D,xie2017rethinking,diba2019dynamonet} which couple appearance and temporal modeling by means of spatiotemporal kernels. 

\begin{figure*}[t] 
\centering
\centerline{\includegraphics[width=.98\linewidth]{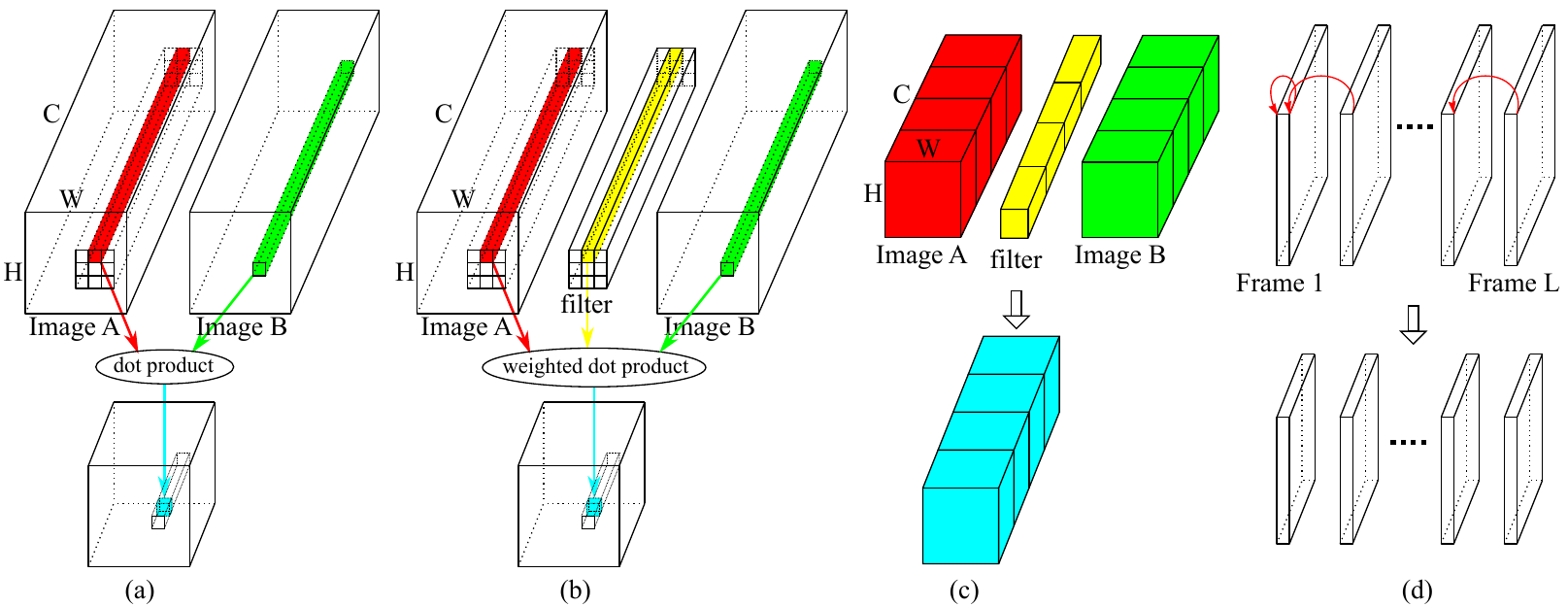}}
   \caption{An illustration of the proposed correlation operator. (a) Correlation operator used for optical flow and geometric matching. (b) The introduction of filters renders the operator ``learnable.'' (c) Groupwise correlation increases the number of output channels without adding computational cost. (d) Extending the correlation operator to work on a sequence of video frames.} 
   \label{fig:corr_op}
\end{figure*}

In this paper we propose a new  scheme based on a novel correlation operator inspired by the correlation layer in FlowNet~\cite{FlowNet}. While in FlowNet the correlation layer is only applied once to convert the video information from the RGB pixel space to the motion displacement space, we propose a learnable correlation operator to establish frame-to-frame matches over convolutional feature maps to capture different notions of similarity in different layers of the network. Similarly to two-stream models, our model enables the fusion of {\em explicit} motion cues with appearance information. However, while in two-stream models the motion and appearance subnets are disjointly learned and fused only in a late layer of the model, our network enables the efficient integration of appearance and motion information throughout the network. Compared to 3D CNNs, which extract spatiotemporal features, our model factorizes the computation of appearance and motion, and learns distinct filters capturing different measures of patch similarity. The learned filters can match pixels moving in different directions.
Through our extensive experiments on four action recognition datasets (Kinetics, Something-Something, Diving48 and Sports1M), we demonstrate that our correlation network compares favorably with widely-used 3D CNNs for video modeling, and achieves competitive results over the prominent two-stream network while being much faster to train. We summarize our contributions as follows:
\begin{myitemize} 
\item A new correlation operator with learnable filters. By making use of dilation and grouping, the operator is highly efficient to compute. Compared to 3D convolution or optical flow, it provides an alternative way to model temporal information in video.
\item A new correlation network which is designed to integrate motion and appearance information in every block. A rigorous study of the new architecture and comparisons with strong baselines provide insights for the different design choices.
\item Our correlation network outperforms the state-of-the-art on four different video datasets without  using optical flow.
\end{myitemize}

In the rest of the paper, we introduce related work in Section~\ref{sec:related}, and detail the proposed correlation operator in Section~\ref{sec:corr_op}. We present the correlation network in Section~\ref{sec:corr_net}. Experimental setups are in Section~\ref{sec:exp_setup}. We discuss the experimental results in Section~\ref{sec:exp_results} and conclude the paper in Section~\ref{sec:conclusion}. 
\section{Related Work}\label{sec:related}

\noindent\textbf{Architectures for video classification.} 
Among the popular video models, there are two major categories: two-stream networks~\cite{SimonyanZ14,wang2015towards,FeichtenhoferPZ16,WangXW0LTG16,qiu2019learning,crasto2019mars} and 3D CNNs~\cite{baccouche2011sequential,Ming2013,Tran15,linsuniccv15,tran2017closer,P3D,xie2017rethinking,diba2019dynamonet,feichtenhofer2020x3d}. 
Since the introduction of
two-stream networks~\cite{SimonyanZ14},
further improvements have been achieved by adding connections between the two streams~\cite{FeichtenhoferPZ16}, or inflating a 2D model to 3D~\cite{I3D}.
3D CNNs~\cite{baccouche2011sequential,Ming2013,Tran15} learn appearance and motion information simultaneously by convolving 3D filters in space and time. Successful image architectures~\cite{SimonyanZ14a,Szegedy15,KaimingHe16} have been extended to video using 3D convolution~\cite{I3D,Tran15,xie2017rethinking}. 
Recent research~\cite{linsuniccv15,tran2017closer,P3D,qiu2019learning} shows that decomposing 3D convolution into 2D spatial convolution and 1D temporal convolution leads to better performance.
Our correlation network goes beyond two-stream networks and 3D convolution, and we propose a new operator that can better learn the temporal dynamics of video sequences.

\begin{table*}[t]
\centering
\begin{tabular}{c|c|c} \hline
	Operator & Correlation & 3D convolution  \\ \shline 
	Input &  $C_{in} \times L \times H \times W$ & $C_{in} \times L \times H \times W$ \\ \hline 
	Filter &  $L \times C_{in} \times K \times K $ & $C_{out} \times C_{in} \times K_{t} \times K_{y} \times K_{x}$  \\ \hline
	Output & $ (G * K * K) \times L \times H \times W$ &  $C_{out} \times L \times H \times W$  \\ \hline 
	\# params & $L * C_{in} * K * K$ & $C_{out} * C_{in} * K_{t} * K_{y} * K_{x}$ \\ \hline
	FLOPs & $C_{in} * K * K * L * H * W$ & $C_{out} * C_{in}  * K_{t} * K_{y} * K_{x} * L * H * W$ \\ \hline
\end{tabular}
\caption{A comparison of the correlation operator with 3D convolution. When the size $K$ of the filter is similar (\ie, $K * K \approx K_{t} * K_{y} * K_{x}$), the parameters of 3D convolution is about $C_{out}/L$ times more than the correlation operator, and its FLOPs is about $C_{out}$ times higher.} 
\label{tab:corr_vs_3d}
\end{table*}

\noindent\textbf{Motion information for action recognition.} 
Before the popularity of deep learning, various video features~\cite{Laptev03,Scovanner07,KMS08,Piotr05,Wang2013} were hand-designed to encode motion information in video.
Besides two-stream networks and 3D CNNs, ActionFlowNet~\cite{ng2018actionflownet} proposes to jointly estimate optical flow and recognize actions in one network. 
Fan~\etal~\cite{fan2018end} and Piergiovanni~\etal~\cite{piergiovanni2018representation} also introduced networks to learn optical flow end-to-end for action recognition. 

There is also work~\cite{sun2018optical,lee2018motion,hommos2018using} seeking alternatives to optical flow. Sun~\etal~\cite{sun2018optical} extracted features guided by optical flow to capture the transformation between adjacent frames. Lee~\etal~\cite{lee2018motion} designed motion filters by computing the difference of adjacent frames. 
Hommos~\etal~\cite{hommos2018using} proposed to use phase instead of optical flow as the motion representation for action recognition. Our paper is along the line of designing architectures to directly learn motion information from raw RGB pixels.

\noindent\textbf{Applications of correlation operation.}  Deep matching~\cite{weinzaepfel2013deepflow} computes the correlation of image patches to find dense correspondence to improve optical flow. 
Unlike deep matching using hand-crafted features, FlowNet~\cite{FlowNet} is a network,
where a correlation layer performs multiplicative patch comparisons. 
Correlation layers were also used in other CNN-based optical flow algorithms~\cite{sun2018pwc,ilg2017flownet}. Besides optical flow, Rocco~\etal~\cite{rocco2017convolutional} used it to estimate the geometric transformation of two images, whereas Feichtenhofer~\etal~\cite{feichtenhofer2017detect} applied it to object tracking.

In the context of action recognition, Zhao~\etal~\cite{zhao2018recognize} utilize the correlation layer to compute a cost volume to estimate the displacement map as in optical flow. 
The Spatio-Temporal Channel Correlation Network~\cite{diba2018spatio} adapts the Squeeze-and-Excitation block~\cite{hu2018squeeze} to a ResNeXt~\cite{xie2017aggregated} backbone. The notion of correlation in~\cite{diba2018spatio} refers to the relationship among the spatial and temporal dimensions of the feature maps, which is different from the matching of adjacent frames studied in our work. We compare our results with~\cite{diba2018spatio} in Section~\ref{sec:compare_sota}.

Our paper extends this line of ideas by introducing a learnable operator based on correlation. Instead of trying to explicitly or implicitly estimate optical flow, the correlation operator is used repeatedly in combination with other operators to build a new architecture that can learn appearance and motion information simultaneously and that achieves state of the art accuracy on various video datasets. 
\section{Correlation Operator}\label{sec:corr_op}

This section describes the proposed correlation operator. We start by reviewing the existing correlation operator over image pairs used in optical flow~\cite{FlowNet,sun2018pwc} and geometric matching~\cite{weinzaepfel2013deepflow,rocco2017convolutional}. We then propose to inject filters into the operator to make it learnable. We discuss how to increase the number of output channels while retaining efficiency and low number of parameters by means of a groupwise variant. We finally generalize the operator to work on sequences of video frames.

\noindent\textbf{Correlation operator for matching.} As shown in Figure~\ref{fig:corr_op} (a), each image is represented by a 3D tensor of size $C \times H \times W$, where $C$ is the number of channels and $H \times W$ is the spatial resolution. Given a feature patch $P^B(i,j)$ in image $B$, we compute the similarity of this patch with another patch $P^{A}(i^{'},j^{'})$ in image $A$, where $(i,j)$ is the spatial location of the patch. To make the computation more tractable, the size of the feature patch can be reduced to a single pixel, thus $P^{A}(i^{'},j^{'})$ and $P^B(i,j)$ becomes $C$-dimensional vectors. The similarity is defined as the dot product of the two vectors:
\begin{equation}\begin{aligned}
    S(i, j, i^{'}, j^{'}) = 1/C * \sum \limits_{c=1}^C (P^B_{c}(i,j) * P^{A}_{c}(i^{'},j^{'})),
    \label{eq:dotproduct}
\end{aligned}\end{equation}
where $1/C$ is for normalization. $(i^{'},j^{'})$ is often limited to be within a $K \times K$ neighborhood of $(i,j)$. $K$ is the maximal displacement for patch matching.
Considering all possible locations of $(i,j)$ and $(i^{'},j^{'})$ in Eq.~\ref{eq:dotproduct}, the output $S$ is a tensor of size $K \times K \times H \times W$, where $K \times K$ can be flattened to play the role of channel to generate a 3D feature tensor ($K^{2} \times H \times W$) like the input image. 

\noindent\textbf{Learnable correlation operator.} Computer vision has achieved impressive results by moving from hand-crafted features~\cite{lowe2004distinctive,dalal:inria-00548587} to learnable deep neural networks~\cite{Krizhevsky12,KaimingHe16}. 
The original correlation operator~\cite{FlowNet,sun2018pwc,weinzaepfel2013deepflow,rocco2017convolutional} does not include learnable parameters and thus it is quite limited in terms of the types of representations it can generate. We propose to endow the operator with a learnable filter as shown in Figure~\ref{fig:corr_op} (b). Our motivation is to learn to select informative channels during matching.
To achieve this goal we introduce a weight vector $W_{c}$ to Eq.~\ref{eq:dotproduct} in the dot product computation: $W_{c} * P^{B}_{c}(i,j) * P^{A}_{c}(i^{'},j^{'})$.
The similarity of two feature patches (\ie, $P^B(i,j)$ and $P^{A}(i^{'},j^{'})$) is often related to how close their spatial location is. 
We thus apply different weight vectors $W_{c}$ to different locations in the $K \times K$ neighbor to take into account the spatial distribution of the matching disparity. 
Thus, the size of each filter is $ C  \times K \times K $ as summarized in Table~\ref{tab:corr_vs_3d}. 

$K$ indicates the maximal displacement when matching two patches. Larger valued $K$ can cover larger regions and encode more information. The downside is that the computational cost grows quadratically \wrt $K$. Inspired by the dilated convolution~\cite{yu2015multi}, we propose to perform dilated correlation to handle large displacement without increasing the computational cost. We enlarge the matching region in image $A$ by a dilation factor $D$.
In practice, we set $K=7$ with a dilation factor of $D=2$ to cover a region of $13 \times 13$ pixels. Besides dilation, we also apply the operator at different spatial scales (as discussed in Section~\ref{sec:corr_net}), which is a popular strategy to handle large displacements in optical flow~\cite{ranjan2017optical}. From Figure~\ref{fig:vis_filter}, filters do learn to select discriminative channels as filters from certain channels are more active than the other. Having different weights in the $K \times K$ neighborhood also enables the filter to learn pixel movements in different directions.

\noindent\textbf{Groupwise correlation operator.} The correlation operator converts a feature map from $C \times H \times W$ to $K^{2} \times H \times W$. In popular CNNs, $C$ can be one to two orders of magnitude larger than $K^{2}$. This means that the correlation operator 
may cause a great reduction in the number of channels.
This is not a problem for applications such as optical flow or geometric matching, where the correlation operator is only applied  once. If we  want to design a network based on the correlation operator and apply it repeatedly, it will reduce the dimension of the channels dramatically, and degrade the representation power of the learned features, as shown by the results in Section~\ref{sec:ablation_and_two_stream}.

Similar to ~\cite{guo2019group}, we propose a groupwise version of the correlation operator that avoids shrinking  the number of channels while maintaining efficiency. Groupwise convolution~\cite{Krizhevsky12,xie2017aggregated} was introduced to reduce the computational cost of convolution by constraining each kernel to span a  subset of feature channels.
Here we utilize this idea to increase the number of output channels without increasing the computational cost. For the groupwise correlation operator, all $C$ channels are split into $G$ groups for both input images and filters, and the correlation operation is computed within each group. The outputs of all groups are stacked together as shown in Figure~\ref{fig:corr_op} (c). This increases the number of output channels by a factor of $G$, to a total of $K^{2} G$ channels. The size of each group is $g=C/G$. By choosing the group size properly, we can control the number of channels without additional cost.

\noindent\textbf{From two images to a video clip.} The original correlation operator is designed for matching a pair of images. In this paper, we apply it for video classification where the input is a sequence of $L$ video frames. We extend the operator to video by computing correlation for every pair of adjacent frames of the input sequence. As the number of adjacent frame pairs is $L-1$ (i.e., one fewer than the number of frames), we propose to compute self-correlation for the first frame in addition to the cross-correlation of adjacent frame pairs, shown in Figure~\ref{fig:corr_op} (d). It can keep the length $L$ of the output feature map consistent with the input, and make the correlation operator easier to use when designing new architectures. The gradual change of filters within each column of Figure~\ref{fig:vis_filter} shows filters learn to follow the motion of pixels across frames when extending the correlation operator to a video clip.

Table~\ref{tab:corr_vs_3d} summarizes our final proposed correlation operator and compares it with the standard 3D convolution. Intuitively, 3D convolution seeks to learn both spatial and temporal representation by convolving a 3D filter in space and time. The correlation operator however, is intentionally designed to capture matching information between adjacent frames. The correlation operator provides an alternative way to model temporal information for video classification, and it has much fewer parameters and FLOPs than the popular 3D convolution.

\section{Correlation Network}\label{sec:corr_net}

\newcommand{\blockt}[4]{\multirow{4}{*}{\(\left[\begin{array}{c}\text{{1$\times$1$\times$1}, #2}\\ \text{3$\times$1$\times$1, #3}\\ \text{1$\times$3$\times$3, #2}\\ \text{1$\times$1$\times$1, #1}\end{array}\right]\)$\times$#4}}

\begin{table}[t]
\centering
    \begin{tabular}{c|c|c}
        \hline
        Layers &     R(2+1)D-26  & Output size \\
        \shline
        conv$_1$ &  1$\times$7$\times$7, 64, stride 1,2,2  & L$\times$112$\times$112  \\ 
        \hline
        \multirow{4}{*}{res$_2$} & \blockt{{256}}{{64}}{64}{2} & \multirow{4}{*}{L$\times$56$\times$56} \\
        &  &  \\
        &  &  \\
        &  &  \\
        \hline
        \multirow{4}{*}{res$_3$} & \blockt{{512}}{{128}}{128}{2}  &  \multirow{4}{*}{L$\times$28$\times$28}  \\
        &  &  \\
        &  & \\
        &  & \\
        \hline
        \multirow{4}{*}{res$_4$} & \blockt{{1024}}{{256}}{256}{2}    &  \multirow{4}{*}{$\frac{L}{2}\times$14$\times$14}  \\
        &  & \\
        &  & \\
        &  & \\
        \hline
        \multirow{4}{*}{res$_5$} & \blockt{{2048}}{{512}}{512}{2}  &  \multirow{4}{*}{$\frac{L}{4}\times$7$\times$7} \\
        &  & \\
        &  & \\
        &  & \\
        \hline
        \multicolumn{2}{c|}{global average pool, fc}  & \# classes  \\
        \hline
    \end{tabular} 
\caption{The R(2+1)D backbone for building correlation network.}
\label{tab:arch}
\end{table}

The correlation operator is designed to learn temporal information, and needs to be combined with other operators capturing appearance information in order to yield a comprehensive set of features for video classification.
We first briefly introduce the backbone architecture adapted from R(2+1)D~\cite{tran2017closer}, then discuss how to build the correlation network to leverage the matching information by incorporating the correlation operator into the backbone. 

\noindent\textbf{R(2+1)D backbone.}  The R(2+1)D network~\cite{tran2017closer} was recently introduced and shown to yield state-of-the-art action recognition results on several video datasets. R(2+1)D factorizes the traditional 3D convolution (\ie, $3\times 3 \times 3$)  into a 2D spatial convolution (\ie, $1\times 3 \times 3$) and an 1D temporal convolution (\ie, $3 \times 1 \times 1$).
Decoupling the spatial and temporal filtering is 
beneficial for both hand-crafted features~\cite{Wang2013,Piotr05} and 3D CNNs~\cite{tran2017closer,linsuniccv15,P3D}. Compared with the original R(2+1)D~\cite{tran2017closer}, we make a few changes to further simplify and improve its efficiency, \eg, using bottleneck layers, supporting higher input resolution, keeping the number of channels consistent, less temporal striding, \etc. Table~\ref{tab:arch} provides the details of the R(2+1)D backbone used in this paper.

\begin{figure}
\begin{center}
   \includegraphics[width=0.9\linewidth]{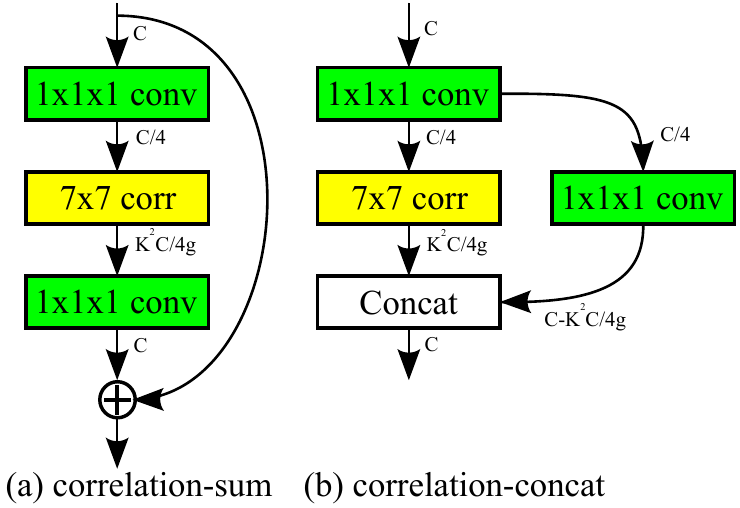}
\end{center}
\caption{Two types of correlation blocks. We mark the number of channels for each operator.} 
\label{fig:corr_block}
\end{figure}

\noindent\textbf{Correlation network.} To incorporate the correlation operator into the backbone network, we propose the two types of correlation blocks shown in Figure~\ref{fig:corr_block}. The design of these blocks is similar in spirit to that of the bottleneck block~\cite{KaimingHe16}. Figure~\ref{fig:corr_block} (a) illustrates the \textit{correlation-sum} block. It first uses an $1\times 1 \times 1$ convolution to reduce the number of channels, then applies a correlation operator for feature matching. Finally another $1\times 1 \times 1$ is used to restore the original number of channels. A shortcut connection~\cite{KaimingHe16} is applied for residual learning. The \textit{correlation-concat} block in Figure~\ref{fig:corr_block} (b) has two branches within the block: one branch with a correlation operator and another branch passing the input feature maps through an $1\times 1 \times 1$.  
The output of the two branches are combined together by concatenation in the channel dimension. We compare the two different designs in Section~\ref{sec:ablation_and_two_stream}.  

We obtain the final correlation network by inserting the correlation block into the R(2+1)D backbone architecture. In this paper, we insert one correlation block after res$_{2}$, res$_{3}$ and res$_{4}$ in Table~\ref{tab:arch}. We omit res$_{5}$ as its spatial resolution is low (\ie, $7 \times 7$). Note that the number of FLOPs of the correlation operator is much lower than 
3D convolution. The correlation network only adds a small overhead to the computational cost of the backbone network. Section~\ref{sec:compare_conv} provides a more quantitative analysis.
\section{ Experimental Setups}\label{sec:exp_setup}

\noindent\textbf{Video Datasets.} We evaluate our model on four video datasets that have rather different properties, emphasizing distinct aspects of action recognition.
{\bf Kinetics}~\cite{kinetics} 
is among the most popular datasets for video classification. It consists of about 300K YouTube videos covering 400 categories. 
{\bf Something-Something}~\cite{goyal2017something} is created by crowdsourcing. This dataset focuses on humans performing predefined basic actions with everyday objects. The same action is performed with different objects (``something") so that models are forced to understand the basic actions instead of recognizing the objects.  It includes about 100K videos covering 174 classes. 
 We note this dataset as Something for short. {\bf Diving48}~\cite{li2018resound} was recently introduced and includes videos from diving competitions. The dataset is designed to reduce the bias of scene and object context in action recognition, and force the model to focus on understanding temporal dynamics of video data. It has a fine-grained taxonomy covering 48 different types of diving with 18K videos in total. The annotations of {\bf Sports1M}~\cite{Karpathy14} are produced automatically by analyzing
the text metadata surrounding the videos. As there are many long videos in Sports1M, we cut them into shorter clips to better utilize the data
and end up with a training set of about 5M samples. For Kinetics and Something, annotations on the testing set are not public available, so we report accuracy on the validation set like others. For Diving48 and Sports1M, we report accuracy on the testing set following the setup by the authors~\cite{li2018resound,Karpathy14}.

\noindent\textbf{Training and Testing.} To train the correlation network, we sample a clip of $L$ (16 or 32) frames with a resolution of $224 \times 224$ from a given video. Some videos in Something do not have enough frames. We simply repeat each frame twice for those videos. 
For data augmentation, we resize the input video to have shorter
side randomly sampled in [256, 320] pixels, following~\cite{wang2017non,SimonyanZ14a}, and apply temporal jittering when sampling clips for training. For the default configuration of our correlation network, we use the \textit{correlation-sum} block, and set the filter size to $K=7$ and group size to $g=32$.
Training is done with synchronous distributed SGD on GPU clusters using Caffe2~\cite{caffe2} with a cosine learning rate schedule~\cite{loshchilov2016sgdr}. We train the model for $250$ epochs in total with the first $40$ epochs for warm-up~\cite{goyal2017accurate} on Kinetics. As Something and Diving48 are smaller datasets, we reduce the training epochs from $250$ to $150$ on them.  For Sports1M, we train $500$ epochs since it is the largest dataset. For testing, we sample $10$ clips uniformly spaced out in the video and average the clip-level predictions to generate the video-level results.  Except in Section~\ref{sec:compare_sota}, all reported results are obtained by training from scratch without pretraining on ImageNet~\cite{deng2009imagenet} or other large-scale video datasets. We only use RGB as the input to our model, unlike two-stream networks~\cite{SimonyanZ14,wang2015towards,FeichtenhoferPZ16,WangXW0LTG16} which use both RGB and optical flow. 
\section{Experimental Evaluation}\label{sec:exp_results}

To demonstrate the advantages of the proposed correlation network, we first compare the correlation operator with temporal convolution in Section~\ref{sec:compare_conv}. We evaluate the correlation network under different  settings to justify our design choices and compare with the two-stream network in Section~\ref{sec:ablation_and_two_stream}. We show that our correlation network outperforms the state of the art on all four datasets in Section~\ref{sec:compare_sota}. Finally, we visualize the learned filters in Section~\ref{sec:visualize}.

\begin{table}[t]
\centering
\resizebox{\columnwidth}{!}{
{\footnotesize
		\begin{tabular}{c|c|c|c|c|c} \hline
			&  & \multicolumn{3}{c}{Top-1 accuracy (\%)} \\
			Model & Length & GFLOPs & Kinetics & Something & Diving \\ \shline
			R2D-26 & 16 & 27.5 & 67.8 & 15.8 & 17.5 \\
			R(2+1)D-26 & 16 & 36.0  & 69.9 & 35.4 & 22.7 \\ 
			CorrNet-26 & 16 & 37.4 & 73.4 & 38.5 & 27.0 \\ \hline
			R2D-26 & 32 & 55.0 & 70.1 & 28.1 & 29.2 \\
			R(2+1)D-26 & 32 & 71.9  & 72.3 & 45.0 & 32.2 \\ 
			CorrNet-26 & 32 & 74.8 & {\bf75.1} & {\bf47.4} & {\bf35.5} \\ \hline
		\end{tabular}
}}
\caption{Correlation networks vs baselines. Our CorrNet significantly outperforms the two baseline architectures on three datasets, at a very small increase in FLOPs compared to R(2+1)D.
Using longer clip length $L$ leads to better accuracy on all three datasets.} 
\label{tab:compare_conv}
\end{table}

\begin{figure}[t] 
\centering
\includegraphics[width=0.85\linewidth]{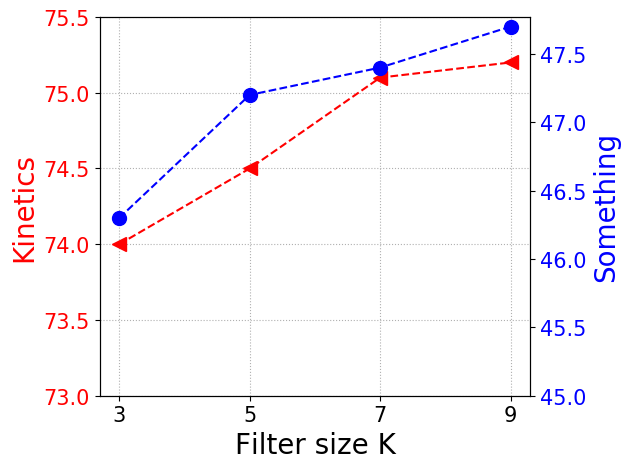}
   \caption{Effect of filter size $K$ on classification accuracy.}
   \label{fig:comp_k}
\end{figure}

\subsection{Correlation network vs baseline backbones} \label{sec:compare_conv}

Table~\ref{tab:compare_conv} compares the correlation network with different baselines. We denote the backbone architecture from Table~\ref{tab:arch} as \textit{R(2+1)D-26}. To demonstrate the importance of temporal learning on different datasets, we create \textit{R2D-26}, which is obtained by removing all 1D temporal convolutions (\ie, $3 \times 1 \times 1$), and adding a $3 \times 1 \times 1$ max pooling when we need to do temporal striding.
 \textit{CorrNet-26} is obtained by inserting one \textit{correlation-sum} block after res$_{2}$, res$_{3}$ and res$_{4}$  of R(2+1)D-26 as described in Section~\ref{sec:corr_net}. As the correlation block adds a small overhead to the FLOPs, we further reduce the number of filters for conv$_1$ from 64 to 32, and remove the $3 \times 1 \times 1$ temporal convolutions from res$_2$ for CorrNet. This reduces the accuracy of CorrNet only slightly (less than $0.5\%$). The resulting CorrNet-26 has similar FLOPs as R(2+1)D-26, as shown in Table~\ref{tab:compare_conv}.

\noindent\textbf{R2D vs R(2+1)D.} The gap between R2D and R(2+1)D varies dramatically on different datasets. On Kinetics and Diving48, R(2+1)D is only 2-5\% better than R2D, but the gap widens up to 20\% on Something. This is consistent with findings in~\cite{xie2017rethinking} and is due to the design of Something where objects are not predictive of the action label. This also highlights the challenges of designing new architectures that can generalize well to different types of datasets.

\noindent\textbf{R(2+1)D vs CorrNet.} We observe a consistent improvement of over 3\% on three datasets when comparing CorrNet with R(2+1)D  in Table~\ref{tab:compare_conv}. We achieve the most significant gain on Diving48, \ie, 4.3\%, using 16 frames.  Note that our improved R(2+1)D is a very strong baseline and its performance is already on par with the best results (listed in Table~\ref{tab:soa_kinetics} and ~\ref{tab:soa_some_dive}) reported. A significant 3\% improvement on three datasets shows the power of the information learned from pixel matching and the general applicability of the correlation network to model video of different characteristics. Moreover, CorrNet only increases the GFLOPs of the network by a very small margin, from 71.9 to 74.8, comparing with R(2+1)D.

\noindent\textbf{Input clip length.} Table~\ref{tab:compare_conv} also compares different models using different input length $L$. As expected, increasing $L$ from 16 to 32 frames can boost the performance across all datasets. Something and Diving48 benefit more from using longer inputs. 
It is noteworthy that the improvements of CorrNet over R(2+1)D are largely carried over when using 32 frames. To simplify, we use $L=32$ frames in all the following experiments.

\begin{table}[t]
\centering
    \begin{tabular}{c|c|c}
        \hline
        Datasets &   Kinetics  &  Something \\
        \shline
        CorrNet-26 & 75.1 & 47.4 \\\hline \hline
        w/o filter & 73.9 & 46.5 \\
        w/o grouping & 74.2 & 46.1 \\
        \textit{correlation-concat}  & 73.2 & 45.9 \\\hline
    \end{tabular} 
\caption{Action recognition accuracy (\%) for different configurations of CorrNet.}
\label{tab:comp_param}
\end{table}

\begin{table}[t]
\centering
    \begin{tabular}{c|c|c}
        \hline
       Datasets  &   Kinetics  &  Something \\
        \shline
        CorrNet-26 & 75.1 & 47.4 \\\hline \hline
        R(2+1)D-26 (RGB) & 72.3 & 45.0 \\
        R(2+1)D-26 (OF) & 66.5 & 42.5  \\
       R(2+1)D-26 (Two-stream)  & 74.4 & 47.9 \\ \hline
    \end{tabular} 
\caption{Action recognition accuracy (\%) of CorrNet vs two-stream network.}
\label{tab:comp_two_stream}
\end{table}

\subsection{Evaluating design choices and comparison to two-stream network} \label{sec:ablation_and_two_stream}

To justify our design choices, we experimentally compare different configurations of CorrNet-26 in Table~\ref{tab:comp_param}. We consider the following modifications: 
1) remove filters in the correlation operator; 2) remove grouping to reduce the number of channels from $C$ to $K^{2}$; 3) swap the \textit{correlation-sum} block with \textit{correlation-concat}. Note that we only change one thing at a time.

Removing filters results in an accuracy drop of 1\% on both datasets, as it significantly reduces the power of the learned representations. Similarly, the aggressive channel reduction introduced by removing grouping also causes an accuracy drop of about 1\%. 
The \textit{correlation-concat} block performs worse than \textit{correlation-sum}, which leverages the shortcut connection to ease optimization.

Figure~\ref{fig:comp_k} shows the performance of CorrNet-26 for $K\in\{3, 5, 7, 9\}$. As expected, a larger $K$ can cover a larger neighborhood while matching pixels, thus yields a higher accuracy. But the improvements become marginal beyond $K=7$, possibly due to the low resolution of the feature maps.

\begin{table} [t] 
\centering
\resizebox{\columnwidth}{!}{%
{\footnotesize
		\begin{tabular}{ccccc} \hline
			 \multirow{2}{*}{\bf Methods}  & \multirow{2}{*}{\bf Pretrain} & {\bf Two} & {\bf GFLOPs} & \multirow{2}{*}{\bf Kinetics}  \\ 
			 & & {\bf stream} & {\bf $\times$ crops} &  \\ \shline 
			STC-ResNext-101~\cite{diba2018spatio} & \redxmark & \redxmark & N/A & 68.7  \\
			R(2+1)D~\cite{tran2017closer} & \redxmark & \redxmark & 152$\times$115 & 72.0  \\
			MARS+RGB~\cite{crasto2019mars} & \redxmark & \redxmark & N/A &  74.8   \\
			ip-CSN-152~\cite{tran2019video} & 	 \redxmark & \redxmark & 109$\times$30 & 77.8 \\ 
			DynamoNet~\cite{diba2019dynamonet}  & 	 \redxmark & \redxmark & N/A & 77.9 \\
			SlowFast-101~\cite{feichtenhofer2019slowfast} & \redxmark & \redxmark  & 213$\times$30 & 78.9 \\
			SlowFast-101+NL~\cite{feichtenhofer2019slowfast} & \redxmark & \redxmark  & 234$\times$30 & 79.8 \\ \hline
			I3D~\cite{I3D} & ImageNet & \redxmark & 108$\times$N/A & 72.1 \\		
			R(2+1)D~\cite{tran2017closer} & Sports1M & \redxmark & 152$\times$115 & 74.3 	 \\
			NL I3D-101~\cite{wang2017non} & ImageNet & \redxmark & 359$\times$30 & 77.7 \\
			ip-CSN-152~\cite{tran2019video} & 	 Sports1M & \redxmark & 109$\times$30 & 79.2 \\
			LGD-3D-101~\cite{qiu2019learning} & ImageNet & \redxmark & N/A & 79.4 \\   \hline					
			R(2+1)D~\cite{tran2017closer} & Sports1M & \greencmark & 304$\times$115 & 75.4 	 \\
			I3D~\cite{I3D} & ImageNet & \greencmark & 216$\times$N/A & 75.7  \\
			S3D-G~\cite{xie2017rethinking}  & ImageNet & \greencmark & 142.8$\times$N/A  & 77.2  \\ 	
			LGD-3D-101~\cite{qiu2019learning} & ImageNet & \greencmark & N/A& 81.2 \\ 		
			\hline \hline 
			{\bf CorrNet-50} & \redxmark & \redxmark & 115$\times$10 & 77.2  \\
			{\bf CorrNet-101} & \redxmark & \redxmark & 187$\times$10 & 78.5  \\ 
			{\bf CorrNet-101} & \redxmark & \redxmark & 224$\times$30 & 79.2 \\
			{\bf CorrNet-101} & Sports1M & \redxmark & 224$\times$30 & {\bf 81.0}   \\  \hline 
		\end{tabular}
}}
	\caption{Compare with the state-of-the-art on Kinetics-400.}
	\label{tab:soa_kinetics}
\end{table}

We compare CorrNet-26 with the two-stream network using the R(2+1)D backbone in Table~\ref{tab:comp_two_stream}. We use the
Farneback~\cite{farneback2003two} algorithm for computing optical flow. The two-stream network of R(2+1)D is implemented by concatenating the features after global average pooling. For R(2+1)D, the accuracy gap between RGB and optical flow is smaller on Something, as Kinetics is arguably more biased towards appearance information. Our CorrNet-26 alone is on par with R(2+1)D-26 using two streams. Note that two-stream network effectively doubles the FLOPs of the backbone and the cost of computing optical flow (not considered here) can be very high as well. This shows that our correlation network is more efficient by learning motion information from RGB pixels directly.

\subsection{Comparison to the state of the art} \label{sec:compare_sota}

The correlation network discussed in the previous sections is based on R(2+1)D-26  with a block configuration of [2, 2, 2, 2] for res$_2$, res$_3$, res$_4$ and res$_5$. To compare with the state-of-the-art, we simply add more layers to the backbone. Following the design of ResNet~\cite{KaimingHe16}, CorrNet-50 uses a block configuration of [3, 4, 6, 3], whereas CorrNet-101 uses [3, 4, 23, 3]. Like in CorrNet-26, a correlation block is inserted after res$_2$, res$_3$ and res$_4$ for CorrNet-50. For CorrNet-101, we insert an additional correlation block in the middle of  res$_4$, so there are 4 correlation blocks in total. Table ~\ref{tab:soa_kinetics}, ~\ref{tab:soa_some_dive} and ~\ref{tab:soa_sports1m} compare the accuracy of CorrNet-50 and CorrNet-101 with several recently published results under different settings. For CorrNet-101 (the last two rows of Table ~\ref{tab:soa_kinetics} and ~\ref{tab:soa_some_dive}) at test time, we sample more clips (30 instead of 10), as done in~\cite{wang2017non,wang2018videos} .

As expected, using deeper models or sampling more clips can further improve the accuracy. Comparing with CorrNet-26 in Table~\ref{tab:compare_conv}, CorrNet-101 is 4.1\%, 4.3\% and 3.1\% better on Kinetics, Something and Diving48, respectively. As Diving48 is the smallest dataset among the four, increasing model capacity may lead to overfitting, thus the improvement is less significant. We also experiment with pre-training CorrNet-101 using the Sports1M dataset~\cite{Karpathy14}. This time we achieve the most significant improvement on Diving48, \ie, 6.1\%. Smaller datasets are likely to benefit more from pre-training, as we have seen in the case of UCF101~\cite{UCF101} and HMDB51~\cite{HMDB51}. On both Kinetics and Something, we observe a modest improvement of 1-2\% by pre-training on Sports1M.

\begin{table} [t]
\centering
\resizebox{\columnwidth}{!}{%
{\footnotesize
		\begin{tabular}{ccccc} \hline
			 \multirow{2}{*}{\bf Methods}  & \multirow{2}{*}{\bf Pretrain} & {\bf Two} & \multirow{2}{*}{\bf Something} &  \multirow{2}{*}{\bf Diving} \\ 
			 & & {\bf stream} &  & \\ \shline 
			R(2+1)D~\cite{tran2017closer} & \redxmark & \redxmark & & 21.4 \\
			TRN~\cite{zhou2018temporal} & \redxmark  & \redxmark & 34.4 &  \\
			MFNet-C101~\cite{lee2018motion} & \redxmark & \redxmark & 43.9 & \\ \hline
			NL I3D-50~\cite{wang2017non} & ImageNet & \redxmark & 44.4  & \\
			R(2+1)D~\cite{tran2017closer} & Sports1M & \redxmark & 45.7 & 28.9	 \\
			NL I3D-50+GCN~\cite{wang2018videos} & ImageNet & \redxmark &  46.1 & \\
			DiMoFs~\cite{bertasius2018learning} & Kinetics & \redxmark & & 31.4 \\
			Attention-LSTM~\cite{kanojia2019attentive} &  ImageNet & \redxmark & & 35.6 \\
			GST-50~\cite{luo2019grouped} & ImageNet & \redxmark &  48.6 & 38.8 \\
			MARS+RGB~\cite{crasto2019mars} & Kinetics & \redxmark & 51.7 &  \\   \hline
			S3D-G~\cite{xie2017rethinking}  & ImageNet & \greencmark &   48.2 &  \\ 	
			TRN~\cite{zhou2018temporal} & ImageNet  & \greencmark & 42.0 & 22.8 \\
			MARS+RGB+Flow~\cite{crasto2019mars} & Kinetics & \greencmark & 53.0 &  \\						
			\hline \hline
			{\bf CorrNet-50} & \redxmark & \redxmark &  49.3  & 37.9 \\
			{\bf CorrNet-101} & \redxmark & \redxmark & 50.9 & 38.2 \\  
			{\bf CorrNet-101} & \redxmark & \redxmark &  51.7 & 38.6 \\
			{\bf CorrNet-101} & Sports1M & \redxmark &  {\bf 53.3} & {\bf 44.7}  \\  \hline 
		\end{tabular}
}}
	\caption{Compare with the state-of-the-art on Something-Something v1 and Diving48.} 
	\label{tab:soa_some_dive}
\end{table}

On Kinetics, CorrNet-101 significantly outperforms the previous models using the same setup (\ie, no pretraining and only using RGB) except for the recently introduced SlowFast network~\cite{feichtenhofer2019slowfast} augmented with non-local network (NL)~\cite{wang2017non}. In fact, compared to SlowFast-101, CorrNet-101 achieves slightly higher accuracy  (79.2\% vs 78.9\%), and it is only 0.6\% lower in accuracy when SlowFast-101 is combined with NL. Comparing with results using pre-training, CorrNet-101 is 1.6\% better than LGD-3D~\cite{qiu2019learning}, \ie, 81.0\% vs 79.4\%. The two-stream LGD-3D improves the accuracy to 81.2\% by extracting the computationally expensive TV-L1 optical flow~\cite{zach2007duality}.

Comparing CorrNet-101 with other approaches trained from scratch in Table~\ref{tab:soa_some_dive},  we observe a significant improvement of 7.8\% on Something (51.7\% for CorrNet-101 vs. 43.9\% for MFNet-C101~\cite{lee2018motion}). On Diving48~\cite{li2018resound}, the improvement is even more substantial, \ie, over 17\% (38.6\% from CorrNet-101 vs. 21.4\% from R(2+1)D). With pre-training, CorrNet-101 is still 1.6\% and 5.9\% better on Something and Diving48. CorrNet-101 even slightly outperforms MARS~\cite{crasto2019mars} augmented with RGB and optical flow streams on Something, \ie, 53.3 vs 53.0.

\begin{table} [t] 
\centering
		\begin{tabular}{cccc} \hline
			{\bf Methods}  & {\bf Pretrain} & {\bf Two stream} & {\bf Sports1M}  \\ \shline 
			C3D~\cite{Tran15} &  \redxmark   &  \redxmark &  61.1 \\
			P3D~\cite{P3D}  &  \redxmark  &  \redxmark &  66.4 \\
			R(2+1)D~\cite{tran2017closer} &  \redxmark  & 	\redxmark	& 73.0 \\
			ip-CSN-152~\cite{tran2019video} &  \redxmark  & \redxmark  & 75.5 \\ \hline
			Conv Pool~\cite{yue2015beyond} &  \redxmark & \greencmark & 71.7 \\
			R(2+1)D~\cite{tran2017closer} &  \redxmark & 	\greencmark	& 73.3 \\			
			\hline \hline
			{\bf CorrNet-101} &  \redxmark  & \redxmark &  77.1  \\  \hline 
		\end{tabular}
	\caption{Comparison with the state-of-the-art on Sports1M.} 
	\label{tab:soa_sports1m}
\end{table}

Table~\ref{tab:soa_sports1m} provides a comparison with the state of the art on Sports1M. We only evaluate our best model CorrNet-101 to limit the training time. All the methods in Table~\ref{tab:soa_sports1m} are trained from scratch since Sports1M is already a very large scale video dataset. Our CorrNet-101 established a new state of the art, \ie 77.1\%, outperforming the very recent ip-CSN-152~\cite{tran2019video} by 1.6\%. CorrNet-101 also significantly outperforms R(2+1)D~\cite{tran2017closer} by 3.8\% which uses both RGB and optical flow.

To sum up, CorrNet is a new versatile backbone that outperforms the state-of-the-art on a wide variety of video datasets. Thanks to the efficient design of the correlation operator and our improved R(2+1)D backbone, the FLOPs of CorrNet is also lower than those of previous models, such as NL I3D~\cite{wang2017non}. FLOPs can further be significantly reduced (\ie, 3x decrease) by sampling fewer clips during testing with only a small drop in accuracy, as shown in the third last row of Table~\ref{tab:soa_kinetics} and~\ref{tab:soa_some_dive}.

\begin{figure}
\begin{center}
   \includegraphics[width=0.99\linewidth]{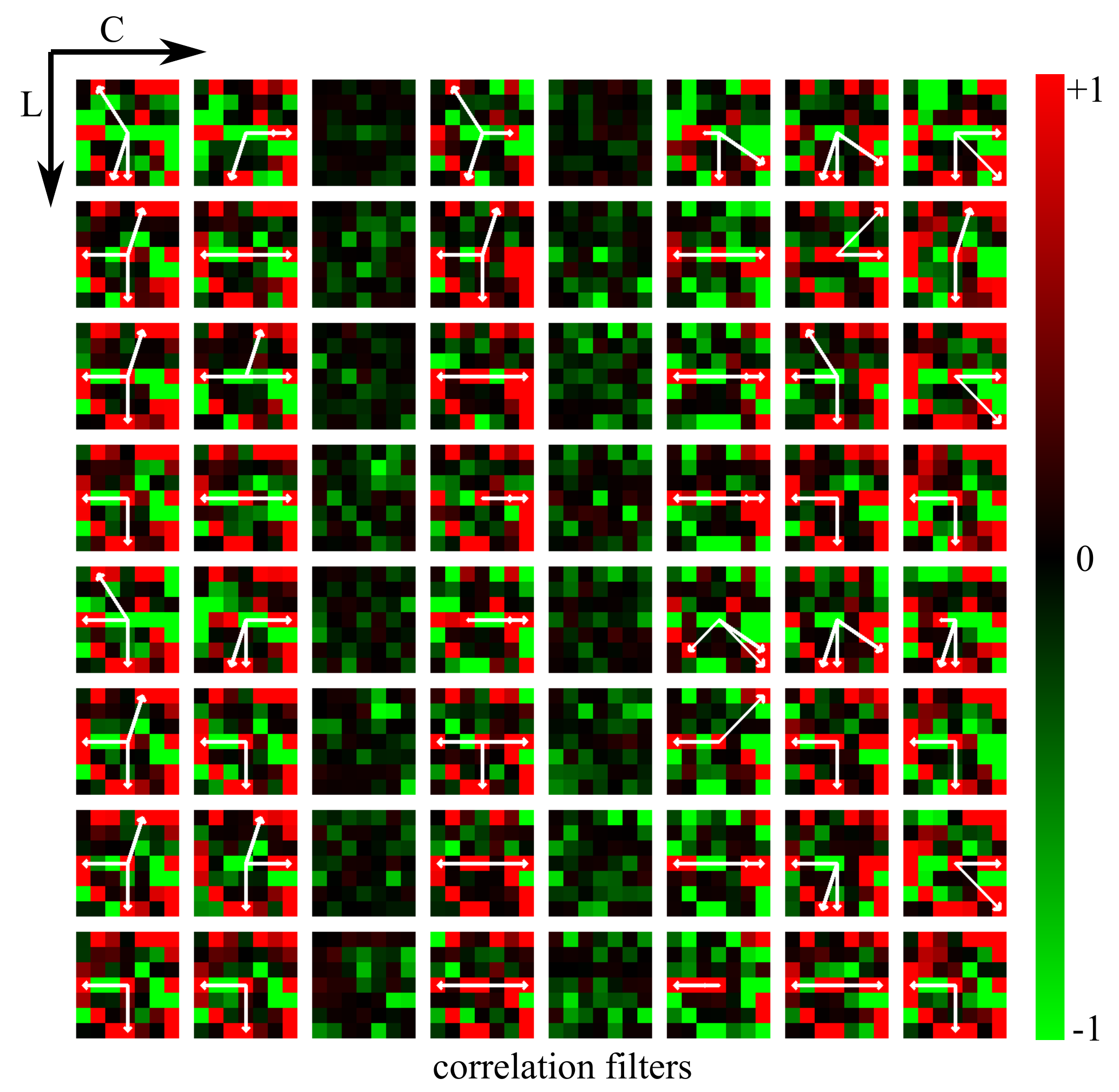}
\end{center}
   \caption{Visualization of CorrNet-101 trained on Kinetics. We visualize the correlation filters, which is a 4D tensor of shape $L \times C \times K \times K$. Filters in each column are aligned in time, and each column represents a different channel dimension. White arrows point to locations with highest weights, showing that different filters learn to match pixels moving in different directions.}
\label{fig:vis_filter}
\end{figure}

\subsection{Visualizing Correlation Filters} \label{sec:visualize}
In this section, we visualize the filters (\ie, the yellow tensor in Fig.~\ref{fig:corr_op}) from the correlation operator to better understand the model. We choose the CorrNet-101 trained from scratch on Kinetics from Table~\ref{tab:soa_kinetics}, and the correlation operator with the highest output resolution, \ie, from the correlation block after res$_2$. The size of the filter is $L \times C \times K \times K$ as listed in Table~\ref{tab:corr_vs_3d}, which is $32 \times 64 \times 7 \times 7$ in this case. We visualize filters for $l=0,\ldots,7$ and $c=0,\ldots,7$ in Figure ~\ref{fig:vis_filter}. The color coding indicates the weights in the learned filters, and the white arrows point to the directions with largest weights.  

Zooming into filters in Figure ~\ref{fig:vis_filter}, we observe that each filter learns a specific motion pattern (\ie, the $7\times7$ grid) for matching. The filters in each column are sorted in time and exhibit similar patterns. The white arrows often point to similar directions for the filters in the same column. This suggests that our network learns the temporal consistency of motion, \ie, pixels usually move in the same direction across frames. Comparing filters in different columns, we observe that some columns are more active than others, which indicates that our filters learns which channels are more discriminative for matching. Filter weights for these channels can be larger than channels that are less informative for matching.
\section{Conclusions}\label{sec:conclusion}

This paper explores a novel way to learn motion information from video data. Unlike previous approaches based on optical flow or 3D convolution, we propose a learnable correlation operator which establishes frame-to-frame matches over convolutional feature maps in the different layers of the network. Differently from the standard 3D convolution, the correlation operator makes the computation of motion information explicit. We design the correlation network based on this novel operator and demonstrate its superior performance on various video datasets for action recognition. Potential future work includes the application of the learnable correlation operator to other tasks, such as action localization, optical flow, and geometry matching.

{\small
\bibliographystyle{ieee_fullname}
\bibliography{ieeedu_ref}
}

\end{document}